\documentclass[a4paper,conference]{IEEEtran}

\ifCLASSINFOpdf
   \usepackage[pdftex]{graphicx}
\else
\fi
\usepackage{algorithmic}

\usepackage{extract}
\usepackage{fancyhdr}
\usepackage{amsfonts}
\usepackage{amssymb}
\usepackage{amsmath}
\usepackage{amsthm}
\usepackage{graphicx}
\usepackage{multirow}
\usepackage{wrapfig}
\usepackage{epsfig}
\usepackage{textcomp}
\usepackage{color}
\usepackage{dirtytalk}
\usepackage{csquotes}
\usepackage{booktabs}
\usepackage{pdflscape}
\usepackage[usenames,dvipsnames]{pstricks}
\definecolor{darkblue}{RGB}{47,68,117}
\definecolor{lightgray}{gray}{0.95}
\usepackage{bbold}
\usepackage{authblk}
\usepackage{float}
\usepackage{caption}
\usepackage{multicol}

\newtheorem*{remark}{Remark}

\newcommand{\be}{\begin{equation}}
\newcommand{\ee}{\end{equation}}
\newcommand{\bea}{\begin{eqnarray}}
\newcommand{\eea}{\end{eqnarray}}
\newcommand{\eqnarr}{\begin{eqnarray}}
\newcommand{\eqnend}{\end{eqnarray}}

\DeclareMathSymbol{\umu}{\mathalpha}{operators}{0}

\newcommand\tab[1][0.cm]{\hspace*{#1}}

\usepackage[linesnumbered,ruled,vlined]{algorithm2e}

\SetCommentSty{mycommfont}
\SetKwInput{KwInput}{Input}                
\SetKwInput{KwOutput}{Output}              

\usepackage{color} 

\usepackage{tikz}
\usetikzlibrary{decorations.pathmorphing} 
\usetikzlibrary{matrix} 
\usetikzlibrary{arrows} 
\usetikzlibrary{calc} 

\tikzstyle{block} = [draw,rectangle,thick,minimum height=1cm,minimum width=3cm, gray, text=black]
\tikzstyle{connector} = [->,thick, gray]
\tikzstyle{line} = [thick, gray]
\tikzstyle{branch} = [circle,inner sep=0pt,minimum size=1.5mm,fill=black,draw=gray]
\tikzstyle{guide} = []
\tikzstyle{snakeline} = [decorate, densely dotted, thick, gray]
\tikzstyle{snakeline*} = [decorate, thick, gray, ->, densely dotted]

\begin{document}
%

\title{Quantifying Model Uncertainty in Inverse Problems via Bayesian Deep Gradient Descent}


\author[1]{Riccardo Barbano\textsuperscript{\textsection}}
\author[1]{Chen Zhang\textsuperscript{\textsection}}
\author[1]{Simon Arridge}
\author[1]{Bangti Jin}
\affil[1]{Department of Computer Science, University College London, Gower Street, London WC1E 6BT, UK}

\maketitle

\begingroup\renewcommand\thefootnote{\textsection}
\footnotetext{Equal contribution. Correspondence to riccardo.barbano.19@ucl.ac.uk.}
\endgroup
\begin{abstract}
Recent advances in reconstruction methods for inverse problems leverage powerful data-driven models, e.g., deep neural networks. These techniques have demonstrated state-of-the-art performances for several imaging tasks, but they often do not provide uncertainty on the obtained reconstruction. In this work, we develop a scalable, data-driven, knowledge-aided computational framework to quantify the model uncertainty via Bayesian neural networks. The approach builds on, and extends deep gradient descent, a recently developed greedy iterative training scheme, and recasts it within a probabilistic framework. Scalability is achieved by being hybrid in the architecture: only the last layer of each block is Bayesian, while the others remain deterministic, and by being greedy in training. The framework is showcased on one representative medical imaging modality, {\it viz.} computed tomography with either sparse view or limited view data, and exhibits competitive performance with respect to state-of-the-art benchmarks, e.g., total variation, deep gradient descent and learned primal-dual.
\end{abstract}


\IEEEpeerreviewmaketitle

\section{Introduction}
The task of reconstructing an unobservable signal or image $x$ from a given collection of observations $y$ that have undergone a corruption process (often following a complex forward transform) is ubiquitous in nearly all scientific disciplines, and represents an integral component of many scientific investigations. One notable feature of such a task is that the underlying problem is often ill-posed, in the sense that small perturbations in the observational data can lead to large deviations in the reconstruction. 

The pipeline of most reconstruction procedures usually begins with deriving an accurate forward model. To cope with the inherent ill-posedness of the inverse problem, one popular idea is to employ regularisation techniques, either explicitly via variational regularisation or implicitly via iterative regularisation (Landweber, Gauss-Newton and Expectation-Maximisation). One prominent class of reconstruction algorithms is based on variational regularisation, which involves minimising a Tikhonov functional that consists of two terms, a fidelity term measuring the fitting quality of the model output to the observational data, and a penalty term, which encodes \emph{a priori} knowledge about the sought-for signal \cite{TikhonovArsenin:1977,EnglHankeNeubauer:1996,ItoJin:2015}. From a statistical standpoint, it may be viewed as maximum \emph{a posteriori} estimation of a certain posterior probability density function. During the past few decades many hand-crafted penalties and prior distributions, such as Sobolev smoothness, sparsity, total (generalised) variation, and anatomical priors, have been proposed, and have achieved impressive performances.
We refer to the class of methods that explicitly defines the forward operator and the probability distribution of the noise as knowledge-driven methods. However, for many applied inverse problems the choice and design of a proper penalty term remain highly nontrivial and an unsuitable choice may greatly compromise the reconstruction accuracy. Furthermore, the mathematical model is often only an approximate description of the real-world physical process, and the high complexity associated with the minimisation problem can present a major computational bottleneck.

In recent years deep learning methods have been widely employed to solve inverse problems 
\cite{sun2016deep, hammernik2018learning, gupta2018cnn,adler2018learned} and offer novel computational frameworks to tackle the aforementioned shortcomings of knowledge-driven approaches. There are several different strategies for using deep neural networks (DNNs) for image reconstruction. For example, to reconstruct signals or images purely based on training data (i.e., approximating the inverse map \cite{Zhu:2018}); to build effective penalty terms directly learned from the training data (or even without training data, e.g., deep image prior \cite{Ulyanov:2018}); and to replace components of an established optimisation algorithm (e.g., gradient descent, proximal gradient iteration, primal-dual algorithm \cite{adler2018learned} and ADMM \cite{sun2016deep}) by DNNs. See the work \cite{arridge2019solving} for a recent overview. Very encouraging empirical results have also been demonstrated for several classical and challenging inverse problems, such as image denoising \cite{zhang2017beyond}, super-resolution \cite{dong2015image}, undersampled MRI \cite{hyun2018deep}, low-dose computed tomography (CT) \cite{yang2018low} and photo-acoustic tomography \cite{hauptmann2018model}. Despite the promising empirical results, purely data-driven approaches have some shortcomings: the need for a large amount of training data (which is infeasible to acquire {in medical} applications), the lack of good interpretability of deep architectures, and the lack of robustness with respect to adversarial attacks. 
The unrolled iteration approach \cite{GregorLecun:2010,putzky2017recurrent, hauptmann2018model,MongaLiEldard:2019} aims at combining the strengths of knowledge-driven and data-driven approaches, by explicitly incorporating the physical models (prior knowledge etc.) and exploiting the good approximation properties and expressivity of DNNs. Thus, this class of methods is very promising for solving many challenging inverse problems, and has received enormous attention recently \cite{ongie2020deep}. 

All the aforementioned works focus on producing one single reconstruction for a given set of corrupted observations, and do not provide relevant uncertainty estimates. Characterising uncertainty in DNN-based solutions to inverse problems is still in its infancy and remains an interesting open problem. In view of the lack of robustness of DNNs, especially in sensitive domains, e.g., medical imaging \cite{antun2020instabilities}, uncertainty estimates provide valuable additional insights. In the context of deep learning there are several different types of uncertainty, e.g., epistemic uncertainty (for instance, the uncertainty associated with the estimation of the parameters of the model), aleatoric uncertainty, and algorithmic uncertainty. The present work focuses on epistemic uncertainty (broadly speaking, also known as model uncertainty), which refers to the fact that for a given training dataset there is a multitude of parameter configurations of the model that can explain the data, each giving a different prediction on the test dataset. In practice, this occurs frequently due to the severe over-parametrisation of DNNs. In this case, it is desirable to assess all the admissible configurations, i.e., placing uncertainty over our model parameters \cite{gal2016uncertainty}. Recovering such uncertainty enables the conveying of additional information on the confidence we have in the model prediction and is thus important for comprehensive down-stream decision making. Uncertainty quantification of inverse problems can be naturally formulated within a Bayesian framework \cite{KaipioSomersalo:2005}. In recent years, uncertainty quantification with deep learning has also been extensively explored, and more recently mined for learned inversion techniques. The work \cite{adler2018deep} proposes a data-driven sampling technique for exploring the posterior distribution of the inverse solution based on Wasserstein generative adversarial networks (GANs). In particular, \cite{zhang2019probabilistic} proposes a probabilistic framework based on conditional variational autoencoders (VAEs), which potentially also allows for incorporating physics, and demonstrates its performance on Gaussian and Poisson denoising. However, both \cite{adler2018deep,zhang2019probabilistic} quantify the uncertainty inherent to the data, i.e., aleatoric uncertainty, but do not provide the uncertainty of the model that fits the data, i.e., epistemic uncertainty.

To quantify the epistemic uncertainty of a learned model, one popular approach is to use Bayesian neural networks (BNNs), and to encode the epistemic uncertainty in the network parameters (conditioned on the training data), through a probability distribution \cite{graves2011practical, blundell2015weight}. However, the exact posterior and the posterior predictive distributions are intractable.
A common method to sample from the posterior is the Markov chain Monte Carlo (MCMC) \cite{gamerman2006markov}, which although asymptotically exact, often does not scale well on BNNs \cite{papamarkou2019}. Moreover, MCMC still cannot be routinely applied to inverse problems in imaging due to high parameter dimensionality. Note that there is important recent progress in this direction using tools from convex optimisation for the posterior distribution on the inverse solution \cite{Pereyra:2019}. 

In the literature, variational inference (VI) \cite{blei2017variational, zhang2018advances} is often adopted as a practical approximate inference scheme for BNNs. Standard BNN implementations tend to double the number of parameters per layer \cite{blundell2015weight}, and consequently, VI often exhibits slow convergence, which can potentially greatly compromise the performance \cite{Osawa:2019, rossi2019good}. There has been extensive research into designing practical approximate inference methods that allow easily scaling up to deep neural architectures. In particular, the connections between VI and dropout \cite{srivastava2014dropout} have been the subject of research, and led to approaches, which are referred to as Monte Carlo dropout (MCDO) schemes \cite{gal2016dropout, gal2016uncertainty}. 
They have shown promising empirical results in several applications but their drawbacks and poor performances were also noted in sequential decision problems \cite{osband2016risk, foong2019pathologies}.

It is still of great importance to develop a scalable inference procedure providing epistemic uncertainty for learning-based inversion techniques. In this work we propose an efficient, data-driven, knowledge-aided computational framework for quantifying epistemic uncertainty based on BNNs for unrolling type learned inversion methods. In particular, our method can be viewed as a probabilistic analogue of a recently proposed deep gradient descent \cite{hauptmann2018model}, and hence the framework is termed Bayesian Deep Gradient Descent (BDGD). Our main contributions are summarised as follows:
\begin{itemize}
    \item We introduce a tractable and statistically principled framework that provides epistemic uncertainty. This is achieved by integrating a data-driven knowledge-aided framework with advances in BNNs and VI. 
    \item We propose a ``greedy'' training scheme, which trains the framework block-wise, in a manner similar to deep gradient descent \cite{hauptmann2018model}. This allows for greatly reducing the training time. Further, we provide an interpretation of the procedure as performing a sequence of constrained VI problems.
    \item Building upon \cite{riquelme2018deep} we further achieve computational efficiency for training a hybrid architecture where VI is only applied on a small portion of the whole network, together with the greedy training scheme.
\end{itemize}

The proposed framework is evaluated on CT reconstruction, from either sparse view or limited view data. The numerical results show that the approach is capable of delivering mean estimators that are competitive with benchmark algorithms, while also providing useful uncertainty information.

The remainder of the paper is organised as follows. In Section \ref{sec:method} we introduce the proposed framework and formalise the training and inference procedures. In Section \ref{sec:experiments} we showcase the proposed framework on CT reconstruction.
In Section \ref{sec:conclusion} we provide concluding remarks. 

\section{Bayesian Deep Gradient Descent}\label{sec:method}
\begin{figure*}[h!]
\centering
\includegraphics[width=0.6\textwidth]{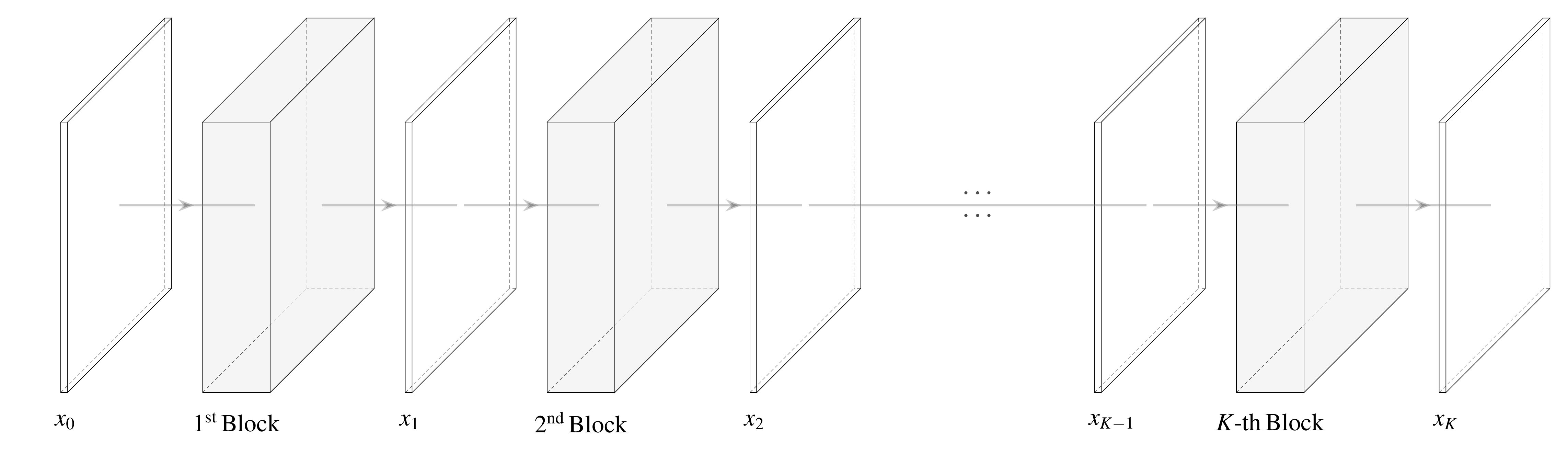}\\
\includegraphics[width=0.6\textwidth]{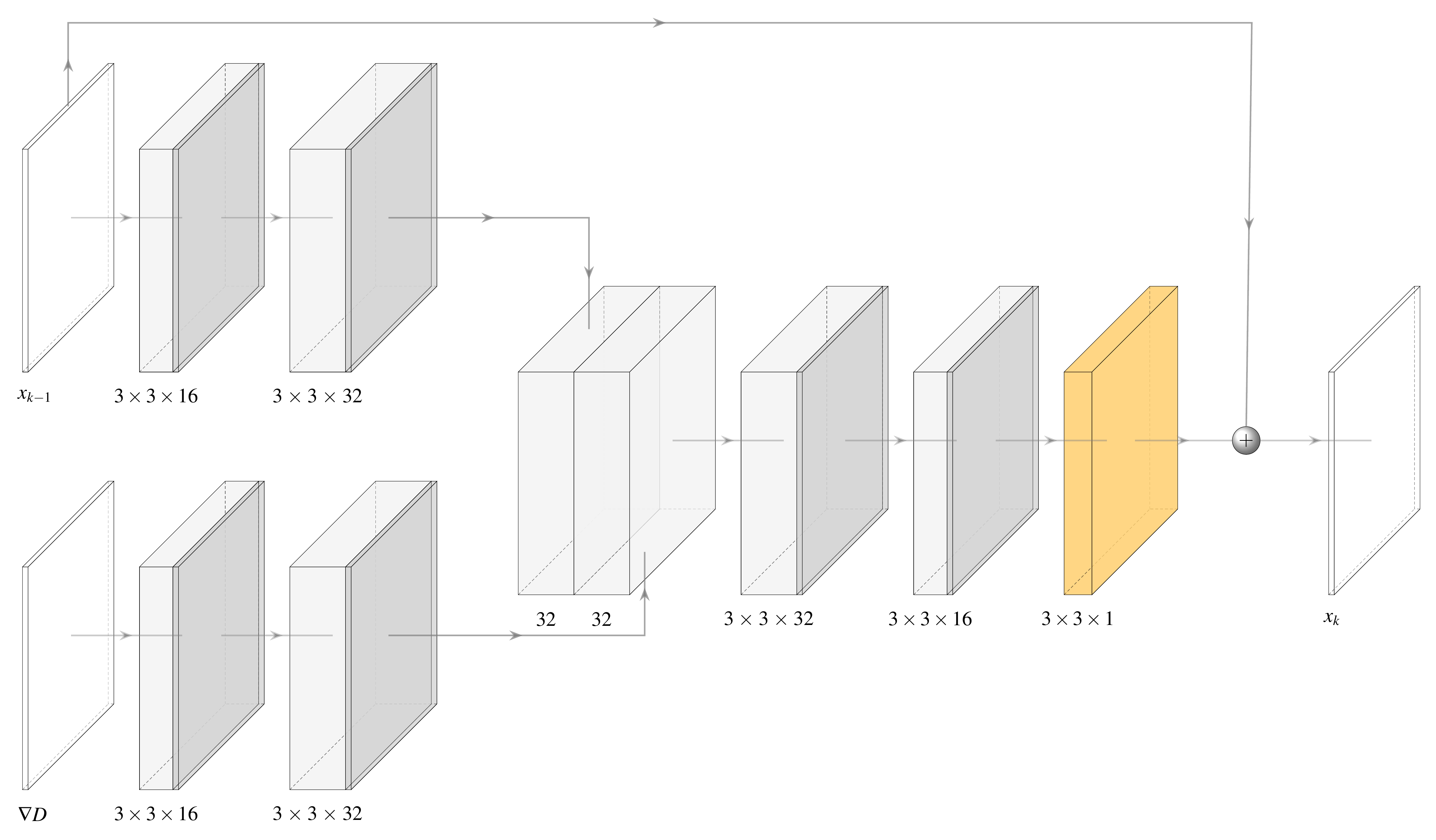}
\caption{ \emph{(Top)} Overall diagram of a $K$-block cascade. \emph{(Bottom)} Diagram of the CNN architecture representing one step update with the $k$-th block. The deterministic layers with parameter $\phi_k$ are colour-coded in gray. The Bayesian convolutional layer with parameter $\theta_k$ is colour-coded in yellow. Note that a ReLU, i.e., the projection, is applied after the summation between the skip connection and the output of the Bayesian convolutional layer.}
\label{fig:architecture}
\end{figure*}

\subsection{Learning Gradient Descent}
In practice, we use iterative methods to optimise the resulting Tikhonov functional. 
The associated minimisation problem for recovering the signal $x$ from the measurements $y$ is given by
\begin{equation}
    x^{\ast} \in \mathrm{argmin}_{x\in\mathcal{C}} \left\{ \mathcal{L}_{\mathrm{MAP}} = D(y, Ax) + \lambda R(x)\right\},
\end{equation}
where $D(y, Ax)$ is a data fidelity term, $A$ the forward operator describing 
the data formation mechanism, $R(x)$ a penalty term expressing \emph{a priori} knowledge about the signal $x$, $\mathcal{C}$ a constraint set, and $\lambda \in \mathbb{R}^{+}$ a regularisation parameter that balances the data fidelity term and the regularisation term $R(x)$. In an unrolled iteration scheme, one constructs a cascade of convolutional neural networks (CNNs) that mimic an iterative minimisation algorithm, and each block within the cascade learns the update for the next iteration:  
\begin{equation}
    x_{k} = f_{\phi_{k}}\left(\nabla D\left(y, Ax_{k-1}\right), x_{k-1}\right), 
\end{equation}
where $f_{\phi_k}$ denotes one block of the cascade and $\phi_k$ is the corresponding parameter vector. Note that $f_{\phi_k}$ can incorporate a feasibility projection operator $P_\mathcal{C}$.
At each iteration, $x_{k-1}$ is updated by the information passed through the gradient $\nabla D\left(y, Ax_{k-1}\right)$. In this way the reconstruction procedure recurrently incorporates the information about $A$ and its adjoint, which often encodes various important physical laws of the inverse problem. Fig. \ref{fig:architecture} shows the network architecture of a single block, how it mimics the gradient update, and the overall cascade. The two branches within each block analyse the information conveyed by $x_{k-1}$ and $\nabla D\left(y, Ax_{k-1}\right)$ separately. The analysis resulting from these two branches is then merged and further processed to give the update $\delta x_{k-1}$.
The update is done by $x_{k} = x_{k-1} + \delta x_{k-1}$, where $\delta x_{k-1}$ is the output of the block before the skip connection and mimics the product of step size and update direction in a gradient descent type algorithm. 
By applying the blocks sequentially for $K$ iterations, the reconstructed image is given by: 
\begin{equation}
    x_{K} = \left( f_{\phi_{K}}\circ f_{\phi_{K-1}} \circ \cdots \circ f_{\phi_{1}}\right) (\nabla D, x_{0}) := f_{\Phi_{K}}(\nabla D, x_{0}), 
    \label{eq:cascade}
\end{equation}
where $x_{0}$ is the initial guess.

\subsection{How to Learn in a Bayesian Framework}

To provide epistemic uncertainty for the reconstructed signal, we use Bayesian convolutional neural networks \cite{blundell2015weight} and learn an approximate posterior distribution $q_{\psi}(\theta)$, parametrised by $\psi$, over each block's parameters $\theta$ via stochastic VI.
For notational simplicity, we drop the subscript $\psi$. Compared to vanilla deterministic DNNs, where the parameters are represented by point estimates, BNNs place prior distributions over the parameters, and obtain a posterior distribution via approximate inference \cite{bishop2006pattern}. Accordingly, we place distributions over the parameters of each block. To distinguish between deterministic and stochastic parameters, below we use $\phi$ and $\theta$, respectively.  

Given a training dataset $\mathcal{D} = \{(x_{i}, y_{i})\}_{i=1}^{N}$, denoted by $\{X,Y\}$, where $Y$ is the observation matrix and $X$ is the ground truth matrix, we train the cascade using a greedy approach, i.e., block by block. One important issue is to interpret the resulting greedy scheme in a statistical context. To this end, we recursively define the prior distribution and likelihood function of $\Theta_k:=(\theta_1,\ldots,\theta_k)$ for training the $k$-th block. Specifically, when training the $k$-th block, we have trained the previous $k-1$ blocks, from $f_{\theta_1}$ to $f_{\theta_{k-1}}$, and computed the optimal approximate posterior distribution $q^*(\Theta_{k-1})$. Then we define the joint prior distribution as $p(\Theta_{k})=q^*(\Theta_{k-1})p(\theta_k|\Theta_{k-1})$, where the (conditional) prior $p(\theta_k|\Theta_{k-1})=\mathcal{N}(0,I)$, and take the likelihood function to be $p(x|y,\Theta_{k})=\mathcal{N}(f_{\Theta_k}(\nabla D, x_{0}),\sigma_k^2 I)$, where $\sigma^2_k$ is an additional trainable parameter. The variational family $\mathcal{Q}_k$ for approximating the true posterior distribution $p(\Theta_k|X,Y)$ is chosen to be of the form $q(\Theta_k)=q^*(\Theta_{k-1})q(\theta_k|\Theta_{k-1})$, where $q(\theta_k|\Theta_{k-1})$ is a mean field Gaussian distribution on $\theta_k$. Thus, $q(\theta_{k}|\Theta_{k-1}) = \prod^{D}_{d=1} \mathcal{N}(\mu_{k, d}, \sigma_{k, d}^2) $, where $\psi_{k}=\{(\mu_{k,d}, \sigma_{k, d}^2)\}^{D}_{d=1}$ are the variational parameters of the mean field approximation and $D$ is the number of parameters per block. The optimal approximate posterior $q^*(\Theta_k)$ is then learned by minimising the following constrained loss function over $\mathcal{Q}_k$ with respect to the variational parameters
\begin{equation}
\begin{split}
	\mathcal{L}_{k}(p, q)=&-\int q(\Theta_k)\log p(X|Y,\Theta_k)\mathrm{d}\Theta_k\\ &+ \mathrm{KL}(q(\Theta_k)||p(\Theta_k)),
\end{split}
\label{eqn:general_loss}
\end{equation}
where KL($\cdot||\cdot$) is the Kullback\textendash Leibler divergence \cite{KullbackLeibler:1951}. We employ the local reparametrization trick to compute the gradients \cite{kingma2013auto, kingma2015variational}. Minimising Eq. \eqref{eqn:general_loss} is equivalent to the following constrained KL minimising problem
\begin{equation}
\min_{q\in\mathcal{Q}_k}\mathrm{KL}(q(\Theta_k)||p(\Theta_k|X,Y)),  
\end{equation}
where $p(\Theta_k|X,Y)$ is given by Bayes' rule with likelihood function $p(X|Y,\Theta_k)$ and prior distribution $p(\Theta_k)$.
It is worth noting that by its very construction, the approximate posterior distribution automatically admits a factorisation form with respect to the blocks, i.e., $q^*(\Theta_k)=q^*(\theta_1)\prod_{i=2}^k q^*(\theta_i|\Theta_{i-1})$. 

\subsection{Example:  2-block Cascade}

To illustrate the framework we provide a simple example with two blocks. Let $\theta_1$ and $\theta_2$ be the parameters of the first and second blocks, respectively. When training the first block, we minimise the problem
\begin{equation}
\begin{split}
    \mathcal{L}_1(p,q) = &-\int q(\theta_1)\log p(X|Y,\theta_1)\rm{d}\theta_1\\ &+ \mathrm{KL}(q(\theta_1)||p(\theta_1)).
\end{split}
\end{equation}
Here $p(\theta_1)$ is the prior distribution of $\theta_1$, which is usually taken to be a standard Gaussian distribution, and $q(\theta_1)$ is the approximate distribution in the mean field Gaussian family. By optimising $\mathcal{L}_1(p, q)$, we obtain an optimal approximate posterior distribution of $\theta_1$, i.e., $q^*(\theta_1)$, and we then use it to construct the joint prior distribution of $\theta_1, \theta_2$. 
The joint prior distribution of $\theta_1$ and $\theta_2$ is defined by $p(\theta_1, \theta_2)=q^*(\theta_1)p(\theta_2|\theta_1)$, where $p(\theta_2|\theta_1)$ is the standard Gaussian distribution. The optimal approximate joint distribution $q^*(\theta_1,\theta_2)$ is found within the constrained family
\begin{equation}
	\mathcal{Q}_2 = \{q(\theta_1,\theta_2)|q(\theta_1,\theta_2)=q^*(\theta_1)q(\theta_2|\theta_1)\},
\end{equation}
by minimising the corresponding loss
\begin{equation}
\begin{split}
	\mathcal{L}_2(p, q) = &- \int q^{\ast}(\theta_1)q(\theta_2|\theta_1)\log p(X|Y,\theta_1, \theta_2)\mathrm{d}\theta_1\mathrm{d}\theta_2 \\&+ \mathbb{E}_{q^*(\theta_1)}[\mathrm{KL}(q(\theta_2|\theta_1)||p(\theta_2|\theta_1))],
\end{split}
\end{equation}
equivalently
\begin{equation}
	q^*(\theta_1,\theta_2) = \arg\min_{q\in\mathcal{Q}_2}\mathrm{KL}(q(\theta_1,\theta_2)||p(\theta_1,\theta_2|X,Y)).
\end{equation}
Note that with a sample of $\theta_1\sim q^*(\theta_1)$ and a sample of $\theta_2\sim q^*(\theta_2|\theta_1)$, the composition of the two blocks $f_{\Theta_2}(\nabla D, x_0)$ outputs the mean of $p(x|y,\theta_1, \theta_2)$.

\subsection{Practicalities in Training and Inference}

Generally, VI methods need some tuning to perform well, especially on CNNs; otherwise they may exhibit slow convergence, which can significantly compromise accuracy \cite{khan2018fast}. Therefore, in the proposed framework we use a composition of maps $f_{\theta_k}\circ f_{\phi_k}$ to model the $k$-th block. Specifically, in the $k$-th block, we denote the parameters of the last layer by $\theta_k$, which is a random variable, and parameters of remaining layers in the block by $\phi_k$, which is regarded as a deterministic variable, c.f., Fig. \ref{fig:architecture} (\emph{Bottom}). It is worth noting that we perform VI on the parameters $\theta_k$'s in a greedy manner, and we are still optimising with respect to $\phi_k$'s. In doing so, we regard $\phi_k$'s as hyperparameters, use the likelihood functions with hyperparameters $p_{\Phi_{k}}(x|y,\Theta_k)$ and for the $k$-th block solve for the following problem
\begin{equation}
\begin{split}
	\min_{q\in\mathcal{Q}_k, \; \phi_k} \{\mathcal{L}(\phi_{k}, q) &= -\int q_{\Phi_{k}}(\Theta_k)\log p_{\Phi_{k}}(X|Y,\Theta_k)\mathrm{d}\Theta_k\\ &\hspace{1em} + \mathrm{KL}(q_{\Phi_{k}}(\Theta_k)||p_{\Phi_{k}}(\Theta_k))\}.
\end{split}
\end{equation}
Again, for notational simplicity, in this section we omit the notation of $\Phi$ and, for instance, denote $(f_{\theta_k}\circ f_{\phi_k}\circ \cdots \circ f_{\theta_1}\circ f_{\phi_1})(\nabla D, x_0)$ by $f_{\Theta_k}(\nabla D, x_0)$, instead of $f_{ \Phi_{k}, \Theta_{k}}(\nabla D, x_{0})$. Methodologically, this is equivalent to the variational family being a delta approximation (i.e., mean field with zero variance) on some parameters, but a Gaussian mean field approximation on the remaining ones. Naturally, the hybrid approach can greatly reduce the number of variational parameters, especially if the Bayesian component is only a small portion of each block, and the resulting cascade has an overall complexity comparable with its deterministic counterpart. The resulting construction retains the Bayesian strength for quantifying epistemic uncertainty, while only slightly increasing the computational efforts and memory requirements.

Once all the blocks in the cascade have been trained, the cascade can be used for inference. Each sampling step amounts to a feed forward propagation through the framework, which is computationally very efficient (especially when compared with classical iterative reconstruction algorithms). Recall that the likelihood function of $\Theta_{K}$ is $p(x|y,\Theta_{K}) = \mathcal{N}(f_{\Theta_{K}}(\nabla D, x_0),\sigma^2_{K}I)$, and the approximate posterior distribution is given by $q^*(\Theta_{K})=q^*(\theta_1)\prod_{k=2}^{K} q^*(\theta_k|\Theta_{k-1})$. One can use Monte Carlo (MC) estimators to estimate the statistics of the distribution
\begin{equation}
    q^*(x|y) = \int p(x|y,\Theta_{K})q^*(\Theta_{K})\mathrm{d}\Theta_{K}.
\end{equation}
Specifically, $\mathbb{E}_{q^*(x|y)}[x]$ can be estimated with an unbiased empirical estimator
\begin{equation}
    \hat{\mathbb{E}}[x]:=\dfrac{1}{T}\sum_{t=1}^{T}f_{\hat{\Theta}^{(t)}_{K}}(\nabla D, x_0)\xrightarrow[\;\;T\;\;\xrightarrow{}\;\;\infty\;\;]{}
    \mathbb{E}_{q^*(x|y)}[x],
\end{equation}
with $T$ samples of $\hat{\Theta}_K$ from $q^*(\Theta_{K})$, i.e., $\{\hat{\Theta}^{(t)}_{K}\}_{t=1}^T$. Moreover, the predictive uncertainty of $q^*(x|y)$ can be estimated by
\begin{equation}
   \widehat{\mathrm{Cov}}[x] := \sigma^{2}_K I + \dfrac{1}{T}\sum_{t=1}^{T} f_{\hat{\Theta}^{(t)}_{K}}(\nabla D, x_0)^{\otimes 2} - \hat{\mathbb{E}}[x]^{\otimes 2},
\end{equation}
where $x^{\otimes 2} = xx^{\top}$. Indeed, 
\begin{align*}
\begin{aligned}
    &\mathbb{E}_{q*(x|y)}[x]
    =\int xq^{\ast}(x|y)\mathrm{d}x\\
    &\hspace{1em}= \int \int x\mathcal{N}(f_{\Theta_K}(\nabla D, x_0), \sigma^{2}_K I)q^{\ast}(\Theta_K) \mathrm{d}\Theta_K\mathrm{d}x\\
    &\hspace{1em}= \int \left(\int x\mathcal{N}(f_{\Theta_K}(\nabla D, x_0), \sigma_{K}^{2}I) \mathrm{d}x\right)q^{\ast}(\Theta_K)\mathrm{d}\Theta_K\\
    &\hspace{1em}= \int f_{\Theta_K}(\nabla D, x_0) q^{\ast}(\Theta_K)\mathrm{d}\Theta_K,
\end{aligned}
\end{align*}
and
\begin{align*}
\begin{aligned} 
&\mathbb{E}_{q^{*}\left(x|y\right)}\left[x^{\otimes 2}\right]\\
&\hspace{1em}=\int\left(\int x^{\otimes 2} \mathcal{N}(f_{\Theta_K}(\nabla D, x_0), \sigma^{2}_K I)\mathrm{d} x\right) q^{*}(\Theta_K) \mathrm{d}\Theta_K\\ 
&\hspace{1em}=\int\left(\operatorname{Cov}_{p\left(x | y, \Theta_K\right)}\left[x\right]+\mathbb{E}_{p\left(x | y, \Theta_K\right)}\left[x\right]^{\otimes 2}\right)q^{*}(\Theta_K) \mathrm{d}\Theta_K\\ 
&\hspace{1em}=\int\big(\sigma^{2}_K I + f_{\Theta_K}(\nabla D, x_0)^{\otimes 2}\big) q^{*}(\Theta_K) \mathrm{d}\Theta_K\\
&\hspace{1em}=\sigma^{2}_K I + \int f_{\Theta_K}(\nabla D, x_0)^{\otimes 2} q^{*}(\Theta_K) \mathrm{d}\Theta_K.
\end{aligned}
\end{align*}

It now follows that $\hat{\mathbb{E}}[x]$ and $\widehat{\mathrm{Cov}}[x]$ are unbiased MC estimators of $\mathbb{E}_{q*(x|y)}[x]$ and $\mathrm{Cov}_{q*(x|y)}[x]=\mathbb{E}_{q^{*}\left(x|y\right)}\left[x^{\otimes 2}\right]-\mathbb{E}_{q^{*}\left(x|y\right)}[x]^{\otimes 2}$ with $T$ samples.
The training and inference procedures of our proposed framework are summarised by Algorithms \ref{algo:DBGD_training} and \ref{algo:DBGD_inference}, respectively. For clarity purposes, in both algorithms we reinstate the subscripts $\psi,\Psi$ and $\phi,\Phi$.  

\begin{algorithm}
\DontPrintSemicolon
\small
\KwInput{\small $\#$ reconstruction steps $K$, dataset $\mathcal{D}$, initial guess $x^{(i)}_{0}$, batch-size $M$}  
 
\For{$k$ $\leftarrow$ $\mathrm{1}$ $\mathrm{to\;} K$}{
    Construct network's input:\\
    $\tab[1cm]\mathcal{D}_{k-1} = \{x^{(i)}_{k-1}, \nabla D(y^{(i)}, Ax^{(i)}_{k-1})\}^{N}_{i=1}$\\
    Train the $k$-th network $f_{\phi_{k}, \theta_{k}}(\nabla D(y^{(i)}, Ax^{(i)}_{k-1}), x^{(i)}_{k-1})$:\\
    \CommentSty{$\tab[1cm]$// stochastic mini-batch optimisation}\\
    $\tab[1cm] \psi^{\ast}_{k}, \phi^{\ast}_{k}\leftarrow \arg \min_{q\in \mathcal{Q}_{k}, \phi_{k}}\bigg\{\hat{\mathcal{L}}(\phi_{k}, q)=
    \tab[1cm]-\dfrac{N}{M}\sum_{i}^{M} 
    \mathbb{E}_{\hat{\Theta}_k\sim q_{\Phi_{k}}(\Theta_k)}
    \left[\log p_{\Phi_{k}}(x^{(i)}|y^{(i)},\hat{\Theta}_k)\right] +
    \tab[0.75cm]\mathrm{KL}(q_{\Phi_{k}}(\Theta_k)||p_{\Phi_{k}}(\Theta_k))\bigg\}$\\
    \CommentSty{// update with $\hat{\theta}_k\sim q_{\Phi_{k}}^*(\theta_k|\Theta_{k-1})$}
    $x^{(i)}_{k}\leftarrow f_{\phi_{k},{\hat{\theta}}_{k}}(\nabla D(y^{(i)}, Ax^{(i)}_{k-1}), x^{(i)}_{k-1})$
    }
\KwOutput{\small approximate posterior at each reconstruction step}
\caption{BDGD (Training)}
\label{algo:DBGD_training}
\end{algorithm}

After each block has been trained, we reconstruct the next update $x_{k}$ with 1 MC sample and compute the gradient of the data fidelity term. At the inference stage of our implementation, we use 100 MC samples to estimate the mean image and pixel-wise variance, $\widehat{\mathrm{Var}}[x] = \mathrm{diag}( \widehat{\mathrm{Cov}}[x])$.

\begin{algorithm}
\DontPrintSemicolon
\small
\KwInput{\small $\#$ reconstruction steps $K$, observation $y$, initial guess $x_{0}$, trained parameters ($\Phi_K, \Psi_K$), $\#$ samples $T$}  
{Construct $\nabla D(y, Ax_0)$}\\
\For{$t$ $\leftarrow$ $\mathrm{1}$ $\mathrm{to\;} T$}{
    \CommentSty{//with $\hat{\Theta}_K^{(t)}\sim q_{\Phi_K}^*(\Theta_K)$}\\
    Sample $x^{(t)}_K = f_{\Phi_K, \hat{\Theta}_K^{(t)}}(\nabla D(y, Ax_0), x_0)$\\
    }
Evaluate $\hat{\mathbb{E}}[x]$ and $\widehat{\mathrm{Var}}[x]$ with $\{x^{(t)}_K\}_{t=1}^T$\\
\KwOutput{\small $\hat{\mathbb{E}}[x]$ and $\widehat{\mathrm{Var}}[x]$}
\caption{BDGD (Inference)}
\label{algo:DBGD_inference}
\end{algorithm}

\begin{remark}[On the Approximate Inference's Landscape]
Apart from Mean Field Variational Inference (MFVI), several other Bayesian approximate schemes have been proposed for uncertainty quantification of neural networks, e.g., MCDO \cite{gal2016dropout}. These Bayesian treatments can also be used within the proposed framework as alternatives to MFVI. Note that with other Bayesian treatments, the underlying choice of prior and thus also the approximate posterior is different. We refer to \cite[Section 3.2]{gal2016uncertainty} for the case of MCDO, which has a VI re-interpretation \cite{gal2016uncertainty,gal2016dropout, hron2018variational}. It consists in training a DNN with dropout, and also applying dropout at test time, which can be seen as approximate marginalisation.
\end{remark}

\section{Experimental Results}\label{sec:experiments}
We showcase the performance of BDGD on CT reconstruction, which is one of the most prominent medical imaging modalities. In 2D CT observations are made in the space $\mathcal{Y} \subset H^\frac{1}{2}({\mathbb R} \times [0,\pi))$, which is the range of the forward operator, Radon transform, $R : \mathcal{X} \rightarrow \mathcal{Y}$ with $\mathcal{X} \subset L^2({\mathbb R}^2)$, consisting of line integrals through $\mathcal{X}$ in ray directions $\hat{\omega} \in \mathbb{S}^1$. While the complete data problem in CT is mildly ill-posed and can be exactly reconstructed by filtered back-projection (FBP), in practice often only a subset of data is available, which can be represented by the composition $A = S\circ R$, with $S$ a subsampling of the directions $\Omega \subset [0,\pi)$. We shall consider two different cases of practical interest in the parallel beam geometry, i) sparse view and ii) limited view. 
Moreover, we assume that the projections contain $1\%$ Gaussian noise. The data fidelity term is then given by the standard squared $L^2$ norm, and accordingly, 
\begin{equation}
    \nabla D(y, Ax_{k-1}) = A^{\top}(Ax_{k-1} -y),
\end{equation}
where the notation $A^{\top}$ denotes the (unfiltered) back-projection operator, i.e., the adjoint of $A$. The depth $K$ of the cascade in BDGD is set to either 10 or 20, depending on the problem setting, and the initial guess $x_{0}$ is set to be the FBP solution. 

For benchmarks, we take three approaches: total variation regularisation (TV) \cite{RudinOsherFatemi:1992}, deep gradient descent (DGD) \cite{hauptmann2018model}, and learned primal-dual (LPD) \cite{adler2018learned}. The latter two are well-established deep unrolled iteration approaches. TV reconstruction is computed with the Chambolle-Pock algorithm, with the regularisation parameter selected via grid search. We also report results for a variant of BDGD, which uses MCDO (with dropout rate of 0.1) instead of MFVI. The number of parameters (per block) is 32833, 32978, and 32833 for DGD, BDGD-MFVI and BDGD-MCDO, respectively. Meanwhile, LPD has 253220 shared parameters, which are trained simultaneously instead of greedily.

All the methods are trained on 4000 randomly generated ellipses (all of size $128\times128$), 
with the background having the lowest value 0. We train each block for 150 epochs. BDGD and the benchmarks are all implemented in Python using Operator Discretisation Library (ODL) \cite{adler2017odl}, PyTorch and TensorFlow. To evaluate the operator $A$ and its adjoint, we use the GPU accelerated ASTRA backend \cite{van2016fast}. 

For validation, we construct a second ellipse dataset (Ellipses Phantoms), and also test on the Shepp-Logan (SL) phantom. 
For a quantitative comparison, we use peak signal-to-noise ratio (PSNR), which is computed by averaging the numerical results over five seeds. 

\subsection{Sparse View CT}

\begin{figure}[h!]
\includegraphics[width=\columnwidth]{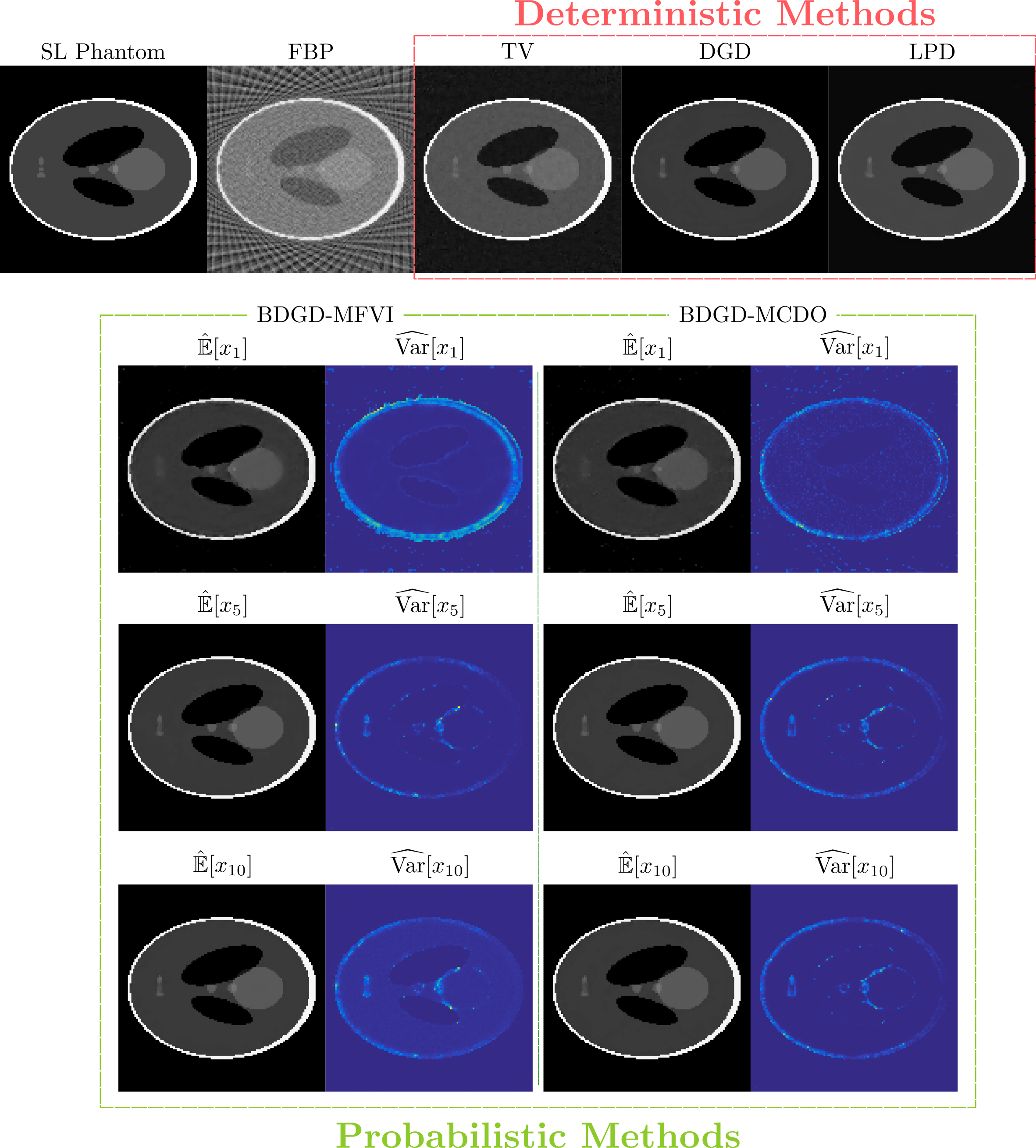}
\caption{Sparse view with 30 directions.}
\label{fig:full_view_sparse_30}
\end{figure}

\begin{figure}[h!]
\includegraphics[width=\columnwidth]{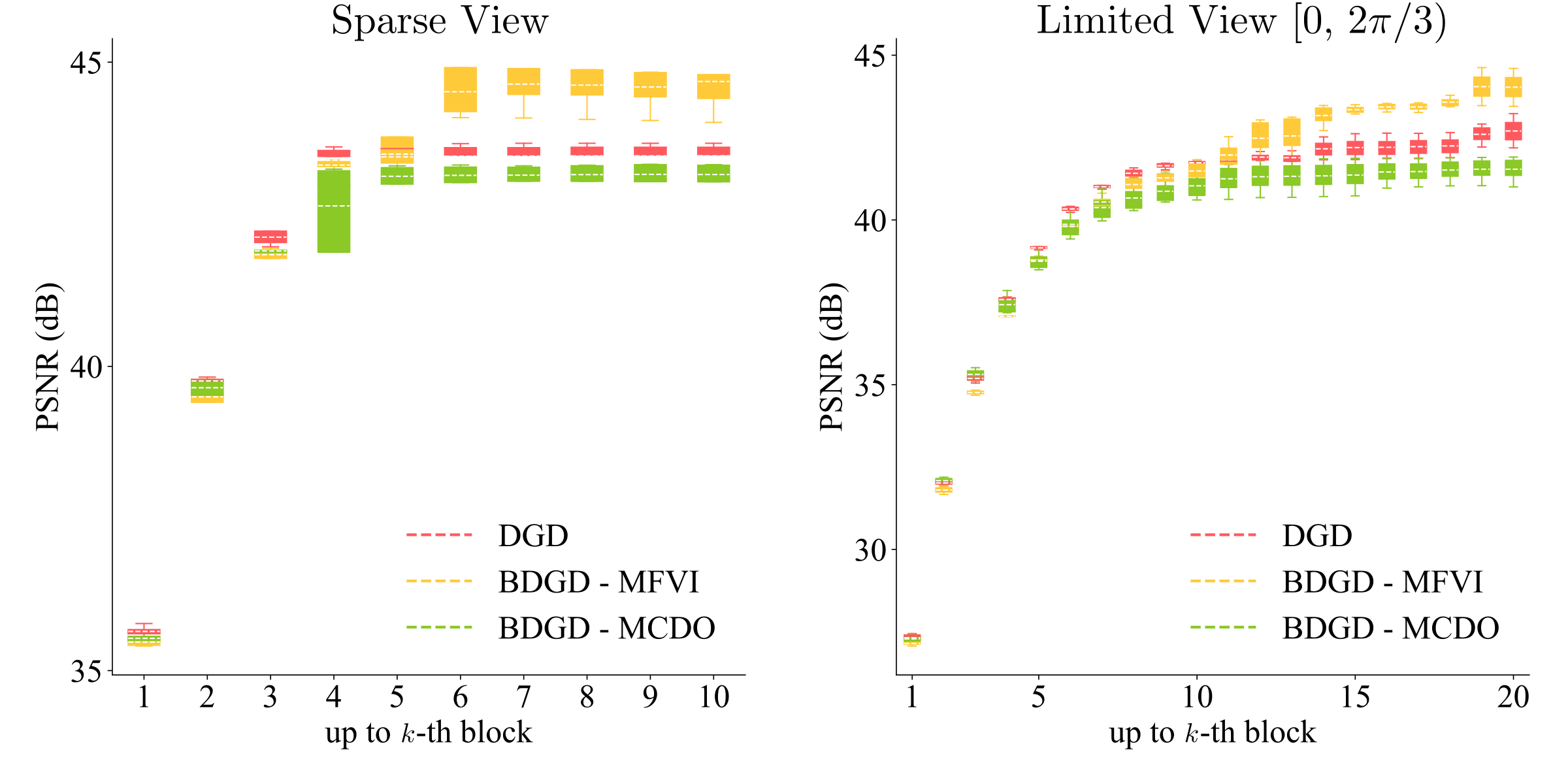}
\caption{PSNR up to $k$-th block: (\emph{Left}) sparse view, (\emph{Right}) limited view $[0, 2\pi/3)$ for the Ellipses Phantoms dataset.}
\label{fig:boxplot}
\end{figure}

In Table \ref{tab:table_full_view_sparse}, we present numerical results for sparse view CT with 30 directions uniformly taken from 0 to $\pi$, computed with $K = 10$ (further increasing the cascade's depth does not lead to better reconstructions). Table \ref{tab:table_full_view_sparse} indicates that BDGD-MFVI outperforms all other methods on both test cases, with the only exception being LPD on the ellipses. In Fig. \ref{fig:full_view_sparse_30}, we show the reconstructions of the SL phantom, and to shed insights into the working mechanisms of BDGD, we show the mean and pixel variance after the first, fifth, and tenth blocks. It is observed that both BDGD-MFVI and BDGD-MCDO show large pixel variance at $k = 1$, which is reduced at $k=5$. However, BDGD-MFVI shows larger pixel variance at $k=10$ than at $k=5$, whereas BDGD-MCDO still presents low epistemic uncertainty. The boxplot in Fig. \ref{fig:boxplot} (\emph{Left}) shows the evolution of PSNR with the number $k$ of blocks. Interestingly, during the first few blocks, BDGD-MFVI and BDGD-MCDO perform comparably, but BDGD-MFVI allows further improvement with additional blocks.

In BDGD-MFVI there appear two distinct phases from a variance standpoint. In the first phase, i.e., the reconstruction phase, the method gradually improves the sample quality and reduces the overall uncertainty. For instance, Fig. \ref{fig:full_view_sparse_30} shows that BDGD-MFVI exhibits high variance up to the first block, i.e., the reconstructions disagree the most. As the method progresses up to the fifth block, the reconstruction shows reduced epistemic uncertainty. In the second phase, i.e., the fine-tuning phase, the method improves the sample quality and gradually reveals more uncertainty for fine details. That is, the greedy training strategy actually incrementally improves the reconstruction. Overall, BDGD-MFVI performs better than BDGD-MCDO, which, in terms of PSNR, behaves nearly identically to the deterministic counterpart (DGD) on the ellipses (we also fail to observe a fine-tuning phase), whereas on the SL phantom, BDGD-MCDO outperforms DGD, showing again the benefit of being Bayesian.

\begin{table}[h!]
\centering
\caption{Sparse View CT}
\resizebox{.8\columnwidth}{!}{%
\begin{tabular}{cccc}
Methods                     &Ellipses Phantoms           &SL Phantom                 \\
\hline\hline 
FBP                         &25.5264                     &18.4667                     \\
TV                          &35.1587                     &37.2162                     \\
LPD                         &\bf{44.5122 $\pm$ 0.4911}   &44.0472 $\pm$ 0.4187        \\
DGD                         &43.2577 $\pm$ 0.4183        &44.6913 $\pm$ 0.6644        \\
BDGD - MFVI                 &\bf{44.6642 $\pm$ 0.4637}   &\bf{47.2946 $\pm$ 0.5778}   \\
\;\;BDGD - MCDO             &43.2126 $\pm$ 0.1285        &45.1725 $\pm$ 0.4461        \\
\hline 
\end{tabular}%
}
\label{tab:table_full_view_sparse}
\end{table}

\subsection{Limited View CT}
We show results for limited view $[0, 2\pi/3)$, using $K=20$; see Table \ref{tab:table_limited_view} for PSNR. As before, BDGD-MFVI outperforms other methods on both datasets in terms of PSNR. Fig. \ref{fig:limited_view_120} shows the reconstruction of an ellipse phantom, along with the pixel variance for BDGD (for both MFVI and MCDO). In the phantom, a larger ellipse almost blends into the background, which is very challenging to reconstruct for all the methods under consideration. However, the presence of the ellipse is well captured by the variance in BDGD-MFVI. Thus, being Bayesian indeed allows a more thorough analysis of the reconstructed image. Note that the uncertainties by BDGD-MFVI capture far more detail about the ellipse edges than those by BDGD-MCDO, and thus are potentially more informative than the latter. Fig. \ref{fig:boxplot} (\emph{Right}) shows the boxplot of the evolution of PSNR with the number of blocks, which exhibits a similar behaviour to the sparse view case.

\begin{center}
\includegraphics[width=.85\columnwidth]{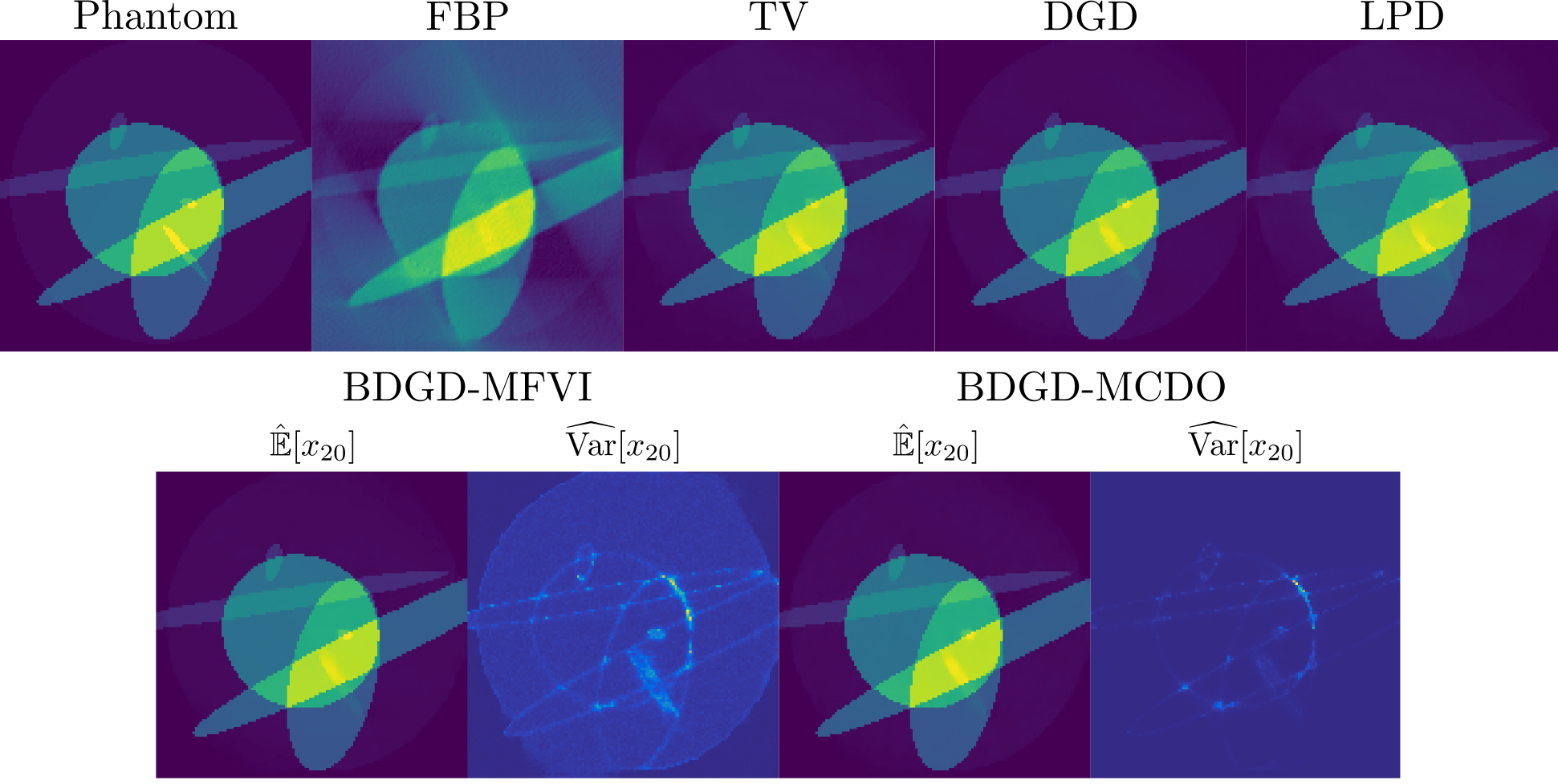}
\captionof{figure}{Limited view $[0, 2\pi/3)$.}
\label{fig:limited_view_120}
\end{center}

\begin{table}[h!]
\centering
\caption{Limited View CT}
\resizebox{.8\columnwidth}{!}{%
\begin{tabular}{ccccc}
Methods                     &Ellipses Phantoms             &SL Phantom                    \\
\hline\hline 
FBP                         &18.5958                        &17.1085                        \\
TV                          &32.9134                        &29.2113                               \\
LPD                         &40.7578 $\pm$ 0.3050           &33.8427 $\pm$ 1.2380                               \\
DGD                         &42.6994 $\pm$ 0.4243           &42.8905 $\pm$ 0.5883           \\
BDGD - MFVI                 &\bf{44.0297} $\pm$ \bf{0.4698} &\bf{45.5140} $\pm$ \bf{0.8261}           \\
\;\;BDGD - MCDO             &41.5367 $\pm$ 0.3884           &41.4397 $\pm$ 0.6299           \\
\hline 
\end{tabular}%
}
\label{tab:table_limited_view}
\end{table}

\subsection{Epistemic Uncertainty with Different Geometries and Unseen Abnormalities}
One of the serious issues with CT reconstructions is the introduction of artefacts due to insufficient information in the data. Thus, it is of significant interest to have indicators on such artefacts. Interestingly, BDGD can provide indicative information by allocating higher variances in related regions, cf. Fig. \ref{fig:understanding_uncertainty_1}, which also includes a more challenging scenario, i.e., limited view $[0, \pi/3)$. In particular, as the view angle range decreases, the magnitude and significant areas of the variance increase, flaring up potential issues with the reconstruction in relevant regions.

\begin{center}
\includegraphics[width=0.75\columnwidth]{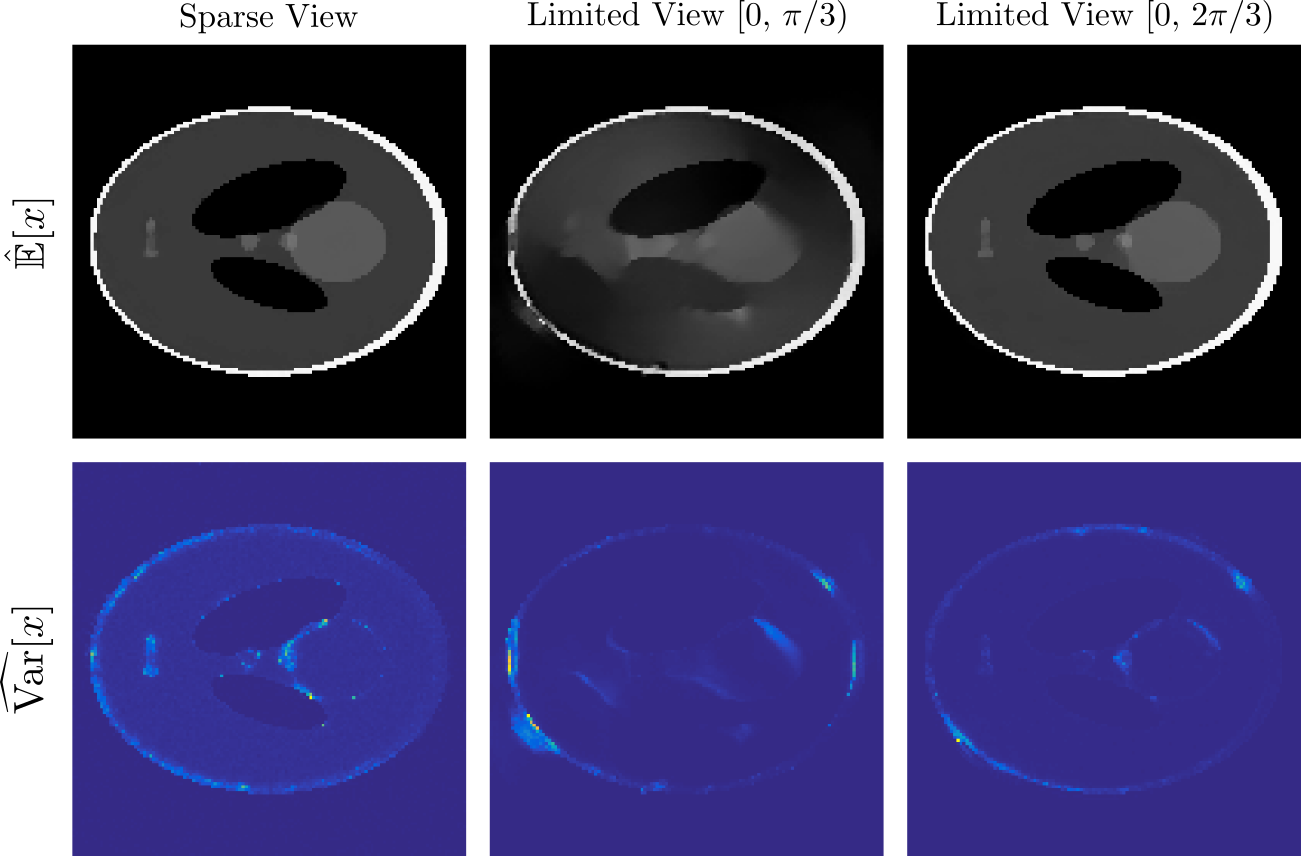}
\captionof{figure}{Mean estimates and epistemic uncertainty maps by BDGD-MFVI for different geometries: (\emph{Left}) sparse view with 30 directions, (\emph{Centre}) limited view $[0, \pi/3)$, (\emph{Right}) limited view $[0, 2\pi/3)$.}
\label{fig:understanding_uncertainty_1}
\end{center}

A very similar phenomenon can be observed, in an even more striking way, in the out-of-distribution test data. In this test, we have added the text ``ICPR 2020" in the ground truth phantom, following the work \cite{antun2020instabilities}. Having abnormal objects in the ellipses patterns, large uncertainty quantities now concentrate on the area around the text, indicating potentially serious issues in the region. This shows the benefit of conveying a notion of uncertainty in addition to a point estimate.

\begin{center}
\includegraphics[width=0.75\columnwidth]{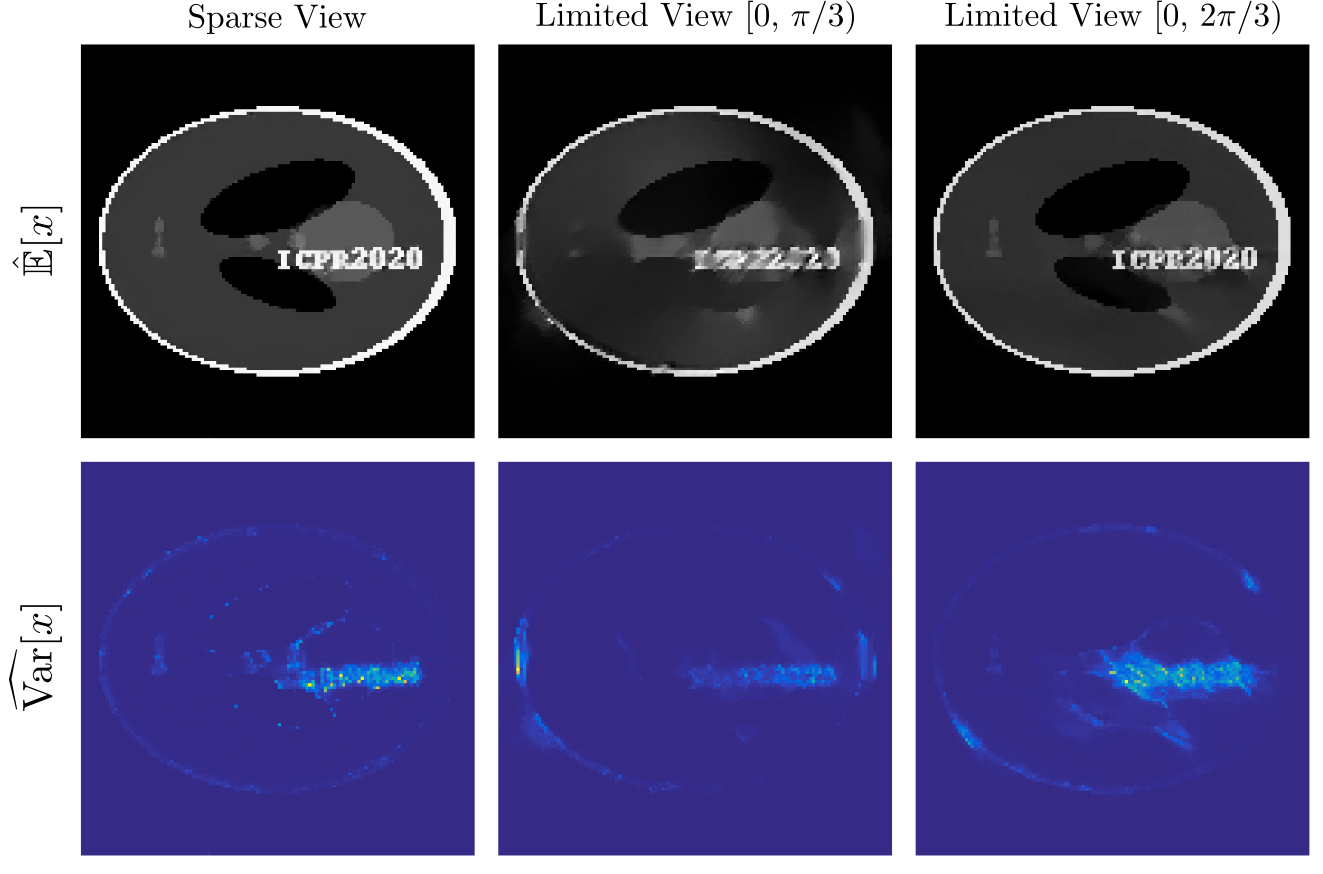}
\captionof{figure}{Out-of-distribution reconstruction for different geometries by BDGD-MFVI: (\emph{Left}) sparse view with 30 directions, (\emph{Centre}) limited view $[0, \pi/3)$, (\emph{Right}) limited view $[0, 2\pi/3)$.}
\label{fig:understanding_uncertainty_2}
\end{center}

\section{Conclusion and Future Work}\label{sec:conclusion}
In this work we have proposed a novel, data-driven, knowledge-aided framework, termed as Bayesian Deep Gradient Descent (BDGD), for providing epistemic uncertainty within the learning based inversion techniques.
We adopted a hybrid model with both point estimation and probabilistic treatment of framework parameters, and realised it using an efficient greedy training strategy in a consistent probabilistic context. The numerical results with CT show that the proposed framework is competitive with state-of-the-art benchmarks. In particular, the results indicate that being Bayesian, even if only a little (one layer per block), can actually be very beneficial to the mean estimate (in terms of PSNR), while also delivering useful uncertainty estimates.
There are several avenues worth pursuing further. First, it is of great interest to explore how the predictive variance information can aid the performance of a downstream processing pipeline, e.g., segmentation of reconstructed images.
Secondly, in view of the outstanding performance of BDGD, it is of much interest to evaluate the significant potential of the framework on more complex medical imaging settings, e.g., photo-acoustic tomography with limited view geometry, and positron emission tomography with a low count emission level. We will explore these possibilities in future works. 

\section*{Acknowledgements}
The authors would like to thank Andreas Hauptmann and Zeljko Kereta for their help in the preparation of the manuscript. RB is supported by a PhD studentship through the EPSRC Centre for Doctoral Training in Intelligent, Integrated Imaging In Healthcare (i4health) (EP/S021930/1), CZ is supported by a UCL Computer Science Departmental Studentship, and BJ is supported by EPSRC EP/T000864/1.

\bibliographystyle{IEEEtran}
\bibliography{bare_conf}

\end{document}